\pdfoutput=1
\documentclass[twoside]{article} 
\usepackage{aistats2018}
\usepackage{latexsym}
\usepackage{pdfpages}
\usepackage{amsmath}
\usepackage{amssymb}
\usepackage{verbatim}
\usepackage{amsfonts}
\usepackage{algorithmicx}
\usepackage[noend]{algpseudocode}
\usepackage{algorithm}
\usepackage{algpseudocode}
\usepackage{multirow}
\usepackage{rotating}
\usepackage[textsize=small]{todonotes}
\usepackage{wrapfig}
\usepackage{epstopdf}
\usepackage{capt-of}
\usepackage{caption}
\usepackage{natbib}
\setcitestyle{round}
\usepackage[inline]{enumitem}
\usepackage{refcount}
\usepackage{dsfont}

\DeclareMathOperator*{\argmax}{arg\,max}
\DeclareMathOperator*{\argmin}{arg\,min}
\algrenewcommand\algorithmicindent{0.5em}%

\newcommand{\mat}[1]{{\pmb #1}}

\makeatletter
\newcommand{\@BIBLABEL}{\@emptybiblabel}
\newcommand{\@emptybiblabel}[1]{}
\makeatother
\usepackage[bookmarks=false]{hyperref}

\begin{document}

\twocolumn[

\aistatstitle{Adversarial Structured Prediction for 
Multivariate 
Measures
}

\aistatsauthor{ Hong Wang \And Ashkan Rezaei \And Brian D. Ziebart} 
\aistatsaddress{Department of Computer Science \\ University of Illinois at Chicago \\ Chicago, IL 60607 \\ \texttt{\{hwang207,arezae4,bziebart\}@uic.edu}}

]

\begin{abstract} 
Many 
predicted structured objects (e.g., sequences, matchings, trees) are 
evaluated using 
the F-score, alignment error rate (AER), or other 
multivariate performance measures.
Since inductively optimizing these measures 
using training data is typically computationally difficult,
empirical risk minimization of surrogate losses is employed, using, e.g.,
the hinge loss for (structured) support vector machines.
These approximations often introduce a mismatch between the learner's objective
and the desired application performance, leading to inconsistency.
We take a different approach: 
adversarially approximate training data 
while optimizing the exact F-score or AER. 
Structured predictions under this formulation result from solving 
zero-sum games between a predictor seeking the best performance and an 
adversary seeking the worst while required to (approximately) match certain 
structured properties of the training data.  
We explore this approach for 
word alignment (AER evaluation) and named entity recognition (F-score 
evaluation) with linear-chain constraints. 
\end{abstract}

\section{INTRODUCTION}
Supervised structured prediction
methods are prevalently used to predict sequences, alignments, or trees,
for 
natural language processing (NLP) tasks~
\citep{manning1999foundations,lafferty2001conditional,och2003systematic,taskar2005discriminative,jurafsky2008speech,finkel2008efficient,haghighi2009better,wang2013joint,durrett2015neural}.
Unfortunately, 
inductively optimizing the
precision, recall, F-score, 
alignment error rate (AER), or 
other multivariate evaluation measures of interest is intractable due to having many
local optima---even 
for the simple 0-1 loss \citep{hoffgen1995robust} and the
structured prediction extension: Hamming loss.


Existing approaches bypass these 
intractabilities 
by replacing the performance measure with a convex
surrogate loss function for which optimization is tractable.  
This can be viewed as
{\bf \emph{approximating the loss function and employing the exact training data}}.
Minimizing the multivariate conditional
logarithmic loss yields  
conditional random fields (CRFs)~\citep{lafferty2001conditional},
and
minimizing the structured hinge loss yields
(structured) support vector
machines (SVM) \citep{tsochantaridis2004support}.
The latter method 
can integrate different multivariate performance measures into the
hinge loss.
However, the mismatch that using a hinge loss approximation introduces
degrades predictive performance in both
theory (i.e., inconsistency \citep{liu2007fisher}) and practice.


We approach the task of structured
predictions with multivariate performance measures
by {\bf \emph{adversarially approximating the training data and 
using the exact evaluation measure}} in two  important NLP tasks:
named entity recognition evaluated using the F-score (\S 2.1) and  
word alignment evaluated using alignment error rate (\S 2.2). 
Our approach (\S\ref{sec:approach}) 
takes the form of a zero-sum game between a predictor player seeking label predictions that minimize expected multivariate loss
and an adversary seeking to approximate the training data labels in
limited ways  based on structural relationships between the variables in a manner that maximizes expected multivariate loss.
This generalizes previous methods for adversarially optimizing multivariate
performance measures without structure \citep{wang2015adversarial} and adversarial structured prediction with chain structures and decomposable losses \citep{li2016adversarial}.
We investigate how these games can be solved efficiently using both exact and approximate constraint generation methods.
Finally, we evaluate our approach in 
\S\ref{sec:experiments_and_results} with comparisons to CRFs and structured SVM methods.

\section{BACKGROUND}

\label{sec:background}

Many important evaluation measures 
are multivariate, meaning they cannot be additively constructed by separately evaluating
on each predicted variable.
We focus on two measures: F-score and alignment error rate (AER) used in
named entity recognition and word alignment tasks. 
We review these tasks before discussing existing 
methods for addressing them. 

\subsection{Named Entity Recognition} 

Identifying all occurrences of a type of entity in a sentence is often needed for natural language processing. Accuracy is often an inappropriate measure due to the relative infrequency of these occurrences.  
Instead, the {\it F-score} is commonly evaluated in 
named entity recognition (NER)~\citep{finkel2005incorporating,tjong2003introduction} and coreference resolution~\citep{raghunathan2010multi,lee2013deterministic} tasks.
It is  the harmonic mean of precision and recall of class $c$ based on  predicted vector $\hat{\bf y}$ evaluated against ground truth vector 
${\bf y}$: 
\begin{align}
F_1^c(\hat{\bf y}, {\bf y}) = \frac{2({\bf 1}_{\hat{\bf y}=c} \cdot {\bf 1}_{{\bf y}=c})}{|{\bf 1}_{\hat{\bf y}=c}|+|{\bf 1}_{{\bf y}=c}|}, \label{eq:f1} 
\end{align}
where 
${\bf 1}_{{\bf y}=c}$ is the binary vector that represents the occurrences of of $c$ in the gold standard sequence. ${\bf 1}_{\hat{\bf y}=c}$ is the binary vector that represents the existence of $c$ in the proposed sequence $\hat{\bf y}$.

Leveraging sequential structure improves performance in named entity 
recognition tasks \citep{finkel2005incorporating}.  This has previously been
accomplished using conditional random fields \citep{lafferty2001conditional} (\S\ref{sec:erm}).
We seek a learning method that is better aligned with the F-score evaluation measure in this paper.



\subsection{Word Alignment}

\begin{figure}[htb]
\centering
\includegraphics[width=0.47\textwidth]{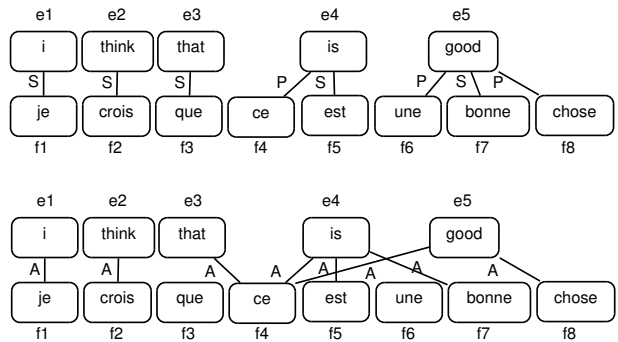}
\caption{The gold standard sequence alignment (top) between English words $e_i$ and French words $f_j$ and the proposed sequence alignment ${\bf a}$ (bottom)
with AER of $\frac{5}{13}$. 
Each complete alignment can be represented in sequence form as
{\it\footnotesize SNNNNNNNN$\hdots$NPSP} and
{\it\footnotesize ANNNNNNNN$\hdots$NNNA} for $(e_1,f_1), (e_1, f_2), \hdots,
(e_1, f_8), (e_2, f_1), \hdots, (e_5, f_8)$ of the gold standard and 
proposed sequence alignments. Negative tags ($N$) are omitted in the figure for clarity.
}
\vspace{-2mm}
\label{fig:aer_example}
\end{figure}

Determining how words align between translated sentences is another important natural language processing task.
The {\it alignment error rate (AER)} is a common evaluation measure for assessing machine translation quality
\citep{cherry2006soft,haghighi2009better,dyer2013simple,kociskylearning}.
It generalizes the F-score for settings with more than binary-valued tags~\citep{och2000improved}.
For example, an alignment task may contain three different kinds of tags between each pair of source and target words: a sure tag ($S$) for unambiguous alignments, a possible tag ($P$) for alignments that might exist or not, and a negative tag ($N$) for alignments that are neither $S$ nor $P$. 
For a gold standard sequence of alignments ${\bf y}$, and a proposed sequence of alignments $\hat{\bf y}$ from the system under evaluation, AER is defined as:
\begin{equation}
\footnotesize
\label{eq:aer_original}
AER(\hat{\bf y}, {\bf y}) = 1 - \frac{{\bf 1}_{\hat{\bf y}=a}\cdot{\bf 1}_{{\bf y}=s} + {\bf 1}_{\hat{\bf y}=a}\cdot{\bf 1}_{{\bf y}=p}}{|{\bf 1}_{\hat{\bf y}=a}| + |{\bf 1}_{{\bf y}=s}|},
\end{equation}
where $A$ is the proposed positive tag. $S$ alignments are also considered to be $P$ alignments ($S \subseteq P$)
in this measure and it is evaluated accordingly. 
Figure \ref{fig:aer_example} shows an example alignment task for an English-French translation.

\subsection{Empirical Risk Minimization Approaches}
\label{sec:erm}

In addition to being inherently multivariate, many key evaluation measures are also not concave (corresponding losses are non-convex).
This makes it challenging to maximize these measures (or perform
empirical risk minimization (ERM) for complementary loss functions):
$\max_\theta \mathbb{E}_{{\bf Y},{\bf X}\sim\tilde{P}, \hat{\bf Y}|{\bf X}\sim \hat{P}_\theta}
\left[ \text{score}(\hat{\bf Y},
{\bf Y})\right]$
for training data distribution $\tilde{P}({\bf y},{\bf x})$ and predictive
distribution $\hat{P}_\theta(\hat{\bf y}|{\bf x})$. 
Even the univariate prediction accuracy (zero-one loss) 
is a non-concave (non-convex) function of
standard prediction model parameters, leading to an
NP-hard optimization problem~\citep{hoffgen1995robust}.
Instead, 
surrogate performance loss measures that upper bound the 0-1 loss/Hamming loss
are employed: logistic regression and conditional random fields (CRFs)~\citep{lafferty2001conditional}---
{\small
\begin{align}
& \hat{P}_\theta({\bf y}|{\bf x}) = \frac{e^{\sum_{t}{\theta} \cdot \Phi({\bf y}_{t-1}, {\bf y}_{t}, {\bf x}_t)}}
{\sum_{{\bf y}}e^{\sum_{t}{\theta} \cdot \Phi({\bf y}_{t-1}, {\bf y}_{t}, {\bf x}_t)}}, \text{ where: } \label{eq:crf}\\
& \theta = \argmin_\theta \mathbb{E}_{{\bf X},{\bf Y}\sim\tilde{P}}
\left[-\log \hat{P}_\theta({\bf Y}|{\bf X})\right] \nonumber
\end{align}
}%
---minimize the logarithmic loss (maximizing log-likelihood); the (structured) support vector machine (SVM)~\citep{cortes1995support,tsochantaridis2004support} and max-margin Markov networks~\citep{tasker2004max} minimize the hinge loss surrogate.
%
The hinge loss
$\xi_i$ grows linearly with the difference between the 
potential $\Phi({\bf x}^{(i)}, {\bf y}^{(i)})$ of the gold standard label 
${\bf y}^{(i)}$ and the best alternative label ${\bf y}'$ when the gold 
standard is not better by at least the multivariate loss between 
${\bf y}'$ and ${\bf y}^{(i)}$, denoted $\Delta({\bf y}',{\bf y}^{(i)})$:
{ \small
\begin{align}
&\min_{\theta, \xi_i \geq 0}
\sum_i \xi_i \label{eq:ssvm} + \frac{\lambda}{2} ||\theta||^2 
\text{ subject to: } \forall i, {\bf y}' \in \boldsymbol{\mathcal{Y}},\\
& {\theta} \cdot [\Phi({\bf x}^{(i)}, {\bf y}^{(i)}) - \Phi({\bf x}^{(i)}, 
{\bf y}')] 
\geq \Delta({\bf y}', {\bf y}^{(i)}) - \xi_i. \nonumber
\end{align}
}%
Though the ability to incorporate multivariate loss functions is attractive,
the hinge loss approximation leads to theoretical shortcomings, including
a lack of Fisher consistency \citep{liu2007fisher}, meaning even if trained
using the true distribution $P({\bf y},{\bf x})$ and an arbitrarily rich
feature representation $\Phi$, predictions minimizing the multivariate loss 
are not guaranteed.

\section{ADVERSARIAL STRUCTURED PREDICTION GAMES}
\label{sec:approach}

We present a generalized adversarial formulation (\S \ref{sec:formulation}) for prediction tasks with both multivariate performance measures and structured properties relating the predicted variables.
We review a general constraint generation method for solving the resulting game problems efficiently (\S\ref{sec:algorithms}) and establish its applicability to the AER performance measure over matchings (\S\ref{sec:aer}), more restrictive bipartite matching (\S\ref{sec:bipartite}),
and F-measure with chain constraints (\S\ref{sec:chainF}).


\subsection{Adversarial Game Formulation}
\label{sec:formulation}

We construct structured predictors for specific performance measures by taking an adversarial philosophy with respect to inductive uncertainty.
For a particular performance measure, 
we seek the predictor that is robust to the worst-case label approximation that still matches structural properties measured on training data or other structural constraints on the predicted variables from the domain.
Mathematically, this takes the form of a zero-sum game
~\citep{neumann1947theory} between two players: player $\hat{\bf Y}$ who maximizes the expected score between the two players' structured choices,
and player $\check{\bf Y}$ who approximates the training data in a manner that minimizes the expected score while also
residing within the constraint set $\Xi$, which is defined as: 
{\small $\mathbb{E}_{{\bf X}\sim\tilde{P},
\check{\bf Y}|{\bf X}\sim \check{P}}
\left[ 
\Phi({\bf X},\check{\bf Y})\right] = \mathbb{E}_{{\bf Y},{\bf X}\sim\tilde{P}}
\left[ 
\Phi({\bf X},{\bf Y})\right]$}: 
{ \small
\begin{align}
& \max_{\hat{P}(\hat{\bf y}|{\bf x})} 
\min_{\check{P}(\check{\bf y}|{\bf x}) \in \Xi}
\mathbb{E}_{{\bf X}\sim\tilde{P},
\hat{\bf Y}|{\bf X}\sim\hat{P},\check{Y}|{\bf X} \sim \check{P}}
\left[\text{score}(\hat{\bf Y}, \check{\bf Y})\right] \label{eq:mpg1} \\
& =
\max_{\theta} \bigg\{
\mathbb{E}_{{\bf Y},{\bf X}\sim \tilde{P}}[\psi({\bf X},{\bf Y})] 
\nonumber\\ & + 
\sum_{{\bf x} \in \boldsymbol{\mathcal{X}}} \tilde{P}({\bf x})   \!\!
\min_{\check{P}(\check{\bf y}|{\bf x})} \max_{\hat{P}(\hat{\bf y}|{\bf x})} 
\Big[
\underbrace{\text{score}(\hat{\bf Y},\check{\bf Y})
-
\psi({\bf x},\check{\bf Y})}_{\mat{S}'_{\hat{\bf y},\check{\bf y}}} 
\Big]\bigg\}, \label{eq:mpg2} 
\end{align}
}where the parameters $\theta$ for the Lagrangian term $\psi({\bf x},\check{\bf y}) = \theta \cdot \Phi({\bf x},\check{\bf y})$, are a learned weight vector, and $\Phi({\bf x},\check{\bf y})$ are the features characterizing the relationship between
${\bf x}$ and $\check{\bf y}$.

\begin{wrapfigure}[10]{r}{0.19\textwidth}
\footnotesize
\vspace{-2mm}
\centering
\includegraphics[width=0.12\textwidth]{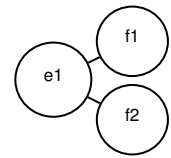}
\caption{Alignment task with one source and two target words.}
\label{fig:aer}
\end{wrapfigure} 
For example, the score matrix (Eq. \eqref{eq:mpg1}) for the $(1-AER)$ game 
for a simple setting with one source
word (e1) and two target words (f1, f2) from Figure \ref{fig:aer} 
is expanded to $\mat{S}'_{\hat{\bf y},\check{\bf y}}$, as shown in Table \ref{tab:aer2},
with Lagrangian terms, $\psi(\check{\bf y})$, 
that enforce the constraint requirement, 
$\check{P}(\check{\bf y}|{\bf x}) \in \Xi$, by motivating the adversary 
to produce features that are similar to those of the training data set.
The predictor chooses (a distribution of) \{A, N\}
tags for the pair of edges, while the adversary chooses (a distribution of) 
\{S, P, N\} tags that also resides within the set $\Xi$.
This corresponds to the game in Eq. \eqref{eq:mpg2}, which can 
be expressed as a linear program.
Estimating game parameters $\theta$ is a convex optimization problem, which 
can be solved using existing convex optimization methods (e.g., AdaDelta,
L-BFGS
). 
%
\begin{table}[htb]
\footnotesize
        \footnotesize
        \centering
\caption{The ($1-AER$) game matrix $\mat{S}'_{\hat{\bf y},\check{\bf y}}$. 
        } 
        \setlength\tabcolsep{3pt}
        \begin{tabular}{|c|c|c|c|c|}
        \hline
         & {\bf NN} & {\bf NA} & {\bf AN} & {\bf AA} \\
        \hline
        {\bf NN}& $1-\psi(NN)$ & $0-\psi(NN)$ & $0-\psi(NN)$ & $0-\psi(NN)$ \\
        \hline
        {\bf NP}& $1-\psi(NP)$ & $1-\psi(NP)$ & $0-\psi(NP)$ & $\frac{1}{2}-\psi(NP)$\\
        \hline
        {\bf NS}& $0-\psi(NS)$ & $1-\psi(NS)$ & $0-\psi(NS)$ & $\frac{2}{3}-\psi(NS)$\\
        \hline
        {\bf PN}& $1-\psi(PN)$ & $0-\psi(PN)$ & $1-\psi(PN)$ & $\frac{1}{2}-\psi(PN)$\\
        \hline
        {\bf PP}& $1-\psi(PP)$ & $1-\psi(PP)$ & $1-\psi(PP)$ & $1-\psi(PP)$\\
        \hline
        {\bf PS}& $0-\psi(PS)$ & $1-\psi(PS)$ & $\frac{1}{2}-\psi(PS)$ & $1-\psi(PS)$\\
        \hline
        {\bf SN}& $0-\psi(SN)$ & $0-\psi(SN)$ & $1-\psi(SN)$ & $\frac{2}{3}-\psi(SN)$\\
        \hline
        {\bf SP}& $0-\psi(SP)$ & $\frac{1}{2}-\psi(SP)$ & $1-\psi(SP)$ & $1-\psi(SP)$\\
        \hline
        {\bf SS}& $0-\psi(SS)$ & $\frac{2}{3}-\psi(SS)$ & $\frac{2}{3}-\psi(SS)$ & 
        $1-\psi(SS)$\\
        \hline
        \end{tabular}
        \label{tab:aer2}
\end{table} 

One key advantage of the adversarial approach over structured prediction
methods based on the hinge loss is its Fisher consistency guarantee 
\citep{li2016adversarial}.  Given the full data distribution,
$P({\bf y},{\bf x})$, and an arbitrarily
rich feature representation, the adversary is constrained by the set
$\Xi$ to exactly match the conditional distribution of the data,
$\check{P}({\bf y}|{\bf x}) = P({\bf y}|{\bf x})$.
Equation \eqref{eq:mpg1} then reduces to:
{ \small
\begin{align}
& \max_{\hat{P}(\hat{\bf y}|{\bf x})} 
\mathbb{E}_{{\bf X}\sim {P},
\hat{\bf Y}|{\bf X}\sim \hat{P}, {\bf Y}|{\bf X} \sim P}
\left[\text{score}(\hat{\bf Y}, {\bf Y})\right],
\end{align}
}%
which is the optimizer of the evaluation measure, $\text{score}(\hat{\bf y},
{\bf y})$, on the full distribution of data.  

\subsection{Double Oracle Method}
\label{sec:algorithms}

Unfortunately, the games for these multivariate settings grow exponentially
in size with the number of variables. 
For example, the $1-AER$ game for the sequence alignment task of
Figure \ref{fig:aer_example} has $5\cdot8$ word pairs, yielding 
$3^{5\cdot8}$ sequence choices for the adversary, $2^{5\cdot8}$ sequence
choices for the predictor player, and a game size of
$(2\cdot3)^{5\cdot8} \approx 1.3\cdot10^{31}$. Thus, explicitly constructing
and solving the game matrix is computationally impractical even 
for small alignment tasks. 


We use the double oracle method~\citep{mcmahan2003planning} to obtain the
equilibrium solution to the adversarial prediction game without explicitly constructing the game matrix.
It iteratively adds a new sequence (for one player at a time) in response to the 
opponent's current equilibrium distribution over sequences obtained from
solving a zero-sum game with the sets of current sequences.
This continues until adding more sequences no longer benefits
either player, guaranteeing a
Nash equilibrium for the original game. 

The double oracle game solver is a central component of the our approach.
Its returned equilibrium distributions
are used to compute the gradients needed for learning 
the model parameters, $\theta$.  
The zero-sum game solver used as a sub-routine of the double oracle method
is implemented using linear programming~\citep{fergusonpart}, 
with different linear program solvers 
available~\citep{optimization2014gurobi,berkelaar2004lpsolve}.
The key remaining problem---also a required subroutine of the double
oracle method---is to efficiently find each player's best response:
{\small
\begin{align}
&\hat{\bf y}^* = \argmax_{\hat{\bf y}} 
\mathbb{E}_{\check{\bf Y}|{\bf x}\sim \check{P}}[
\text{score}(\hat{\bf y},\check{\bf Y}) - \psi(\check{\bf Y})
], 
\label{eq:br1} \\
&
\check{\bf y}^* = \argmin_{\check{\bf y}} 
\mathbb{E}_{\hat{\bf Y}|{\bf x}\sim \hat{P}}[
\text{score}(\hat{\bf Y},\check{\bf y}) - \psi(\check{\bf y})
].
\label{eq:br2}
\end{align}
}%
The difficulty depends on both the loss function being considered and the
constraints imposed on the adversary.
We efficiently solve this best response problem under structured constraints
for alignment error rate and F-score 
with chain structured constraints 
in the next 
sections.

\subsection{Best Response for Alignment Error Rate}
\label{sec:aer}

Under our approach for $1-AER$, the two players are: ${\bf y} \in \mathbb{Y}$, which adversarially approximates the gold standard alignment distribution; and ${\bf a} \in \mathbb{A}$, which maximizes $1-AER$ (hence, minimizes AER), where $\mathbb{A}$ and $\mathbb{Y}$ are the domain of ${\bf a}$ with a distribution $Q(\mathbb{A})$, and ${\bf y}$ with a distribution $Q(\mathbb{Y})$ respectively. For a sequence of alignments of length $n$, the objective of finding the best responses ${\bf a}^*$ (Eq.~\eqref{eq:br1}) and ${\bf y}^*$ (Eq.~\eqref{eq:br2}) 
are:
{ \small
\begin{align}
&{\bf a}^* = \argmax_{{\bf a} \in \mathbb{A}} \!\!\!\!{\sum_{{\bf y} \in {\{0, S, P\}}^n}}\!\!\!\!\!q({\bf y})\left[1-AER({\bf a}, {\bf y}) - \psi({\bf y})\right], \\
&{\bf y}^* = \argmin_{{\bf y} \in \mathbb{Y}} \!\!\!\!{\sum_{{\bf a} \in {\{0, A\}}^n}}\!\!q({\bf a})[1-AER({\bf a}, {\bf y}) - \psi({\bf y})]. \label{eq:aer-br2}
\end{align}
}

We focus first on the adversary's best response (Equation~\eqref{eq:aer-br2}),
which must incorporate the Lagrangian potential term, $\psi({\bf y})$.
For the choice of alignment $y^*$, we separate the choices among all possible numbers of $S$ tags, $k={0, ..., n}$, for the alignment sequence of length $n$, 
and denote these sets as $\mathbb{Y}^{(k)}$. 
The best choice of a certain $k$ is rewritten as follows, where $a_i$ is the notational shorthand of the indicator function $I(a_i = A)$, and $y_{si}=I(y_i = S), y_{ri}=I(y_i=P, y_i\neq S)$; the number of $A$ tags in the alignment is $\alpha = \sum_{i=1}^{n}a_i$:
{\tiny
\begin{align}
& {\bf y}^{(k)^*} = \argmin_{{\bf y} \in \mathbb{Y}^{(k)}} \sum_{{\bf a} \in {\{0, A\}}^n}q({\bf a})\bigg(\frac{\sum_{i=1}^n(2a_iy_{si}+a_iy_{ri})}{\alpha + k} - \sum_{i=1}^{n}\psi_i(y_i)\bigg) \nonumber \\
& = \argmin_{{\bf y} \in \mathbb{Y}^{(k)}} \!\!\!\!\!\!\!
\sum_{\;\;\;{\bf a} \in {\{0, A\}}^n}
\!\!\!\!
q({\bf a})\sum_{i=1}^n\bigg(\frac{a_i(2y_{si}+y_{ri})}{\alpha + k} -\psi_i^{(S)}y_{si} - 
\psi_i^{(R)}y_{ri}\bigg) \label{eq:aer}\\
& = \argmin_{{\bf y} \in \mathbb{Y}^{(k)}} \sum_{i=1}^n y_{si} \underbrace{2\sum_{\alpha=1}^n\left(\frac{q_{i\alpha}}{\alpha+k}-\psi_i^{(S)}\right)}_{f_{sik}}+\sum_{i=1}^n y_{ri} \underbrace{\sum_{\alpha=1}^n\left(\frac{q_{i\alpha}}{\alpha+k}-\psi_i^{(R)}\right)}_{f_{rik}}. \nonumber
\end{align}
}%

Here, in Equation~\eqref{eq:aer}, 
$q_{i\alpha}$ is the marginal probability that alignment $a$ has $|a_S| = \alpha$ and $a_i = S$ (i.e., the number of $S$ tags equals to $\alpha$, and the $i$-th position is $S$).
We separate the Lagrangian potential $\psi$ into two terms: $\psi^{(S)}$ for $S$ tags, $\psi^{(R)}$ for $P$ tags that are not also $S$ tags. 
To get the permutation that minimizes this equation, for tag $S$ at position $i$ we pay $f_{sik}$, for tag $P$ we pay $f_{rik}$, and for tag $N$ we pay $0$. 
Without the ${\bf y} \in \mathbb{Y}^{(k)}$ constraint, to compute the minimum, all that we need to do is finding the smallest of these three terms for each $i$. 
With the constraint, we have to set exactly $k$ tags to $S$, so we choose the $k$ positions where the $f_{sik}$ cost exceeds the best alternative, $\min(f_{rik}, 0)$, by as little as possible.
Thus, we sort $(f_{sik} - \min(f_{rik}, 0))$ in ascending order, set the top $k$ positions to $S$, $P$, or $N$ accordingly. The detailed algorithm is shown as Algorithm~\ref{alg:aer}.

\begin{algorithm}[t]
\caption{AER Maximizer}\label{alg:aer}
\begin{algorithmic}[1]
\Procedure{AerMax}{$Q(\mathbb{A})$, $\psi^{(S)}$, $\psi^{(R)}$}
\State define vector $W_0$ with element $w_{i0}=\frac{1}{i}$
\State define matrix $W$ with element $w_{ik}=\frac{1}{i+k}$, where $i,k\in\{1,...,n\}$
\State compute vector $F_0=QW_0-\psi^{(R)}$
\State compute matrix $F_S=2QW-\psi^{(S)^T}{\bf1}^n$
\State compute matrix $F_R=QW-\psi^{(R)^T}{\bf1}^n$
\State set positions of $y^{(0)^*}$ with $f_{i0} <0$ to `$P$'
\State $\mathbb{E}_{(1-AER)'}({\bf y}^{(0)^*})=\sum_{i=1}^{n}\min(f_{i0},0)$
\For{$k=1$ to $n$}
\State find ${\bf y}^{(k)^*}$ by:
\State sort $f_{ik} = f_{sik} - \min(f_{rik}, 0)$ in asc. order
\State set positions with top $k$'s $f_{ik}$ to `$S$' 
\State set each rest position $i$ to `$P$' if $f_{rik}<0$
\State $\mathbb{E}_{(1-AER)'}({\bf y}^{(k)^*})=\sum_{i=1}^{n}y_{si}f_{sik}+y_{ri}f_{rik}$
\EndFor
\State\Return ${\bf y}^* = \argmin_{{\bf y}^{(k)^*}}{\mathbb{E}_{(1-AER)'}({\bf y}^{(k)^*})}$
\EndProcedure
\end{algorithmic}
\end{algorithm}

The best response for the AER minimizer is simpler to obtain
since the Lagrangian terms $\psi({\bf y})$ are invariant to the choice of 
alignment ${\bf a}^*$.  The approach of
~\citep{dembczynski2011exact} can be used after replacing $F=PW$ with $F'=Q_SW_S+Q_PW_P$, where matrix $Q_S$ is the marginal probability for $S$ tag, $Q_P$ is for $P$ tag, and permutation matrices $W_S$, $W_P$ are for $S$ and $P$ respectively, where each element (with index $i, k$) in the matrix $w_S^{ik} = \frac{2}{i+k}$, and $w_P^{ik} = \frac{1}{i+k-1}$.

\subsection{Best Response for Bipartite Matching}
\label{sec:bipartite}

Following \cite{taskar2005structured}'s modeling of the AER task as a maximum weighted bipartite matching problem using additive cost-sensitive losses~\citep{thomas2001introduction}, we develop a bipartite best response for our adversarial approach.
The best response problems are:
{ \small
\begin{align}
&{\bf a}^* = \argmax_{{\bf a} \in \mathbb{A}} \!\!\!\!{\sum_{{\bf y} \in {\{0, S, P\}}^n}}\!\!\!\!\!q({\bf y})\sum_i \left[1-C({a}_i, {y}_i) - \psi({y}_i)\right], 
\notag \\
&{\bf y}^* = \argmin_{{\bf y} \in \mathbb{Y}} \!\!\!\!{\sum_{{\bf a} \in {\{0, A\}}^n}}\!\!q({\bf a})\sum_i[1-C(a_i, y_i) - \psi(y_i)], \notag 
\end{align}
}%
where each ${\bf a}$ or ${\bf y}$ must also be a valid bipartite matching, and $C(\cdot,\cdot)$ is a cost function.
The best responses can be computed using widely used maximum weight bipartite
matching algorithms~\citep{lawler2001combinatorial} by incorporating Lagrangian potentials and expected losses as edge weights and an integer linear program exactly using the integral solution of a linear program relaxation.


\subsection{Best Response for Linear-chain F-score}
\label{sec:chainF}

\paragraph{Exact Best Responses:}
We consider the F-score for a particular class  $C$ and define two players in the zero-sum game: player $\hat{\bf Y}$ makes predictions that maximizes F-score, and player $\check{\bf Y}$ adversarially approximates the evaluation distribution. For each set of adversarial sequences $\mathbb{Y}$, and its distribution $P(\mathbb{Y})$, the best response should be found efficiently:
{ \small
\begin{align}
&\hat{\bf y}^* = \argmax_{\hat{\bf y} \in \mathbb{\hat{Y}}} {\sum_{\check{\bf y}}} \; p(\check{\bf y})[F_{1_C}(\hat{\bf y}, \check{\bf y}) - \psi(\check{\bf y})], \label{eq:f1_max} \\
&\check{\bf y}^* = \argmin_{\check{\bf y} \in \mathbb{\check{Y}}} {\sum_{\hat{\bf y}}} \; p(\hat{\bf y})[F_{1_C}(\hat{\bf y}, \check{\bf y}) - \psi(\check{\bf y})]. \label{eq:f1_min}
\end{align}
}%

\begin{algorithm}[t]
\caption{Linear-Chain F-score Minimizer}\label{alg:linear_chain_f1}
\begin{algorithmic}[1]
\Procedure{LCFM}{$P(\mathbb{\hat{Y}})$, $\psi^u$, $\psi^b$, $C$}
\State define matrix $W$ with element $w_{tk}=\frac{1}{t+k}$
\State compute matrix $F=2PW$ \Comment element $f_{tk}$
\State $[\check{{\bf y}}^{(0)^*}, \mathbb{E}_{F_1}(\check{{\bf y}}^{(0)^*})]$
$\gets$ \Call{MSUM}{{\it START, 1, 0, 0, F}}
\State $\mathbb{E}_{F_1}(\check{{\bf y}}^{(0)^*}) \gets \mathbb{E}_{F_1}(\check{{\bf y}}^{(0)^*}) - p_0$ \\
\Comment $p_0$ is the marginal probability that $|\hat{y}_C| = 0$ 
\For{$k \gets 1 \dots n$}
\State $[\check{{\bf y}}^{(k)^*}, \mathbb{E}_{F_1}(\check{{\bf y}}^{(k)^*})]\gets$\Call{MSUM}{{\it START,$1,k,k,F$}}
\EndFor
\State $\mathbb{E}_{F_1}(\check{{\bf y}}^{*})\gets \max_{k}(\mathbb{E}_{F_1}(\check{{\bf y}}^{(k)^*}))$
\State $\check{{\bf y}}^{*} \gets \argmax_{\check{{\bf y}}^{(k)^*}}\mathbb{E}_{F_1}(\check{{\bf y}}^{(k)^*})$
\State\Return $[\check{{\bf y}}^{*}, -\mathbb{E}_{F_1}(\check{{\bf y}}^{*})]$
\EndProcedure

\Procedure{MSUM}{$c_{t-1}, t, r, k, F$}
\If{$[\check{{\bf y}}^{*}, \mathbb{E}_{F_1}(\check{{\bf y}}^{*})]$=cache($c_{t-1}, t, r, k$) exists}
\State\Return $[\check{{\bf y}}^{*}, \mathbb{E}_{F_1}(\check{{\bf y}}^{*})]$
\EndIf
\If{$t>n$ and $r>0$} \Return $[\Phi, -\infty]$
\ElsIf{$t>n$ and $r \leq 0$} \Return $[\Phi, 0]$
\EndIf
\State $[\check{{\bf y}}^{*},\mathbb{E}_{F_1}(\check{{\bf y}}^{*})] \gets [\Phi, -\infty]$
\For{$c_t \in \mathbb{C}$}
    \State $\psi(c_t) \gets \psi_t^u(c_t)+\psi_t^b(c_{t-1},c_t)$
    \State $f \gets f_{tk} \times I(c_t = C) \times I(k>0)$
    \If{$r>0 \, |\, c_t \neq C$}
        \State $r' \gets r - I(c_t = C)$
        \State $[\check{{\bf y}}^{c_t},\mathbb{E}_{F_1}(\check{{\bf y}}^{c_t})] \gets $ \Call{MSUM}{$c_t, t+1, r', k, F$}
        \If{$\mathbb{E}_{F_1}(\check{{\bf y}}^{*}) < \psi(c_t)-f+\mathbb{E}_{F_1}(\check{{\bf y}}^{c_t})$}
            \State $\mathbb{E}_{F_1}(\check{{\bf y}}^{*}) \gets \psi(c_t)-f+\mathbb{E}_{F_1}(\check{{\bf y}}^{c_t})$
            \State $\check{{\bf y}}^{*} \gets c_t \oplus \check{{\bf y}}^{c_t}$
        \EndIf
    \EndIf
\EndFor

\State cache($c_{t-1}, t, r, k$) $\gets [\check{{\bf y}}^{*}, \mathbb{E}_{F_1}(\check{{\bf y}}^{*})]$
\State\Return cache($c_{t-1}, t, r, k$) 
\EndProcedure

\end{algorithmic}
\end{algorithm}

Chain structures permit two types of features: unigram $\Phi^u(\mathbf{x}_t,y_t)$, and bigram $\Phi^b(\mathbf{x}_t,y_{t-1},y_t)$.
We encode the linear-chain structure information in the weight 
vectors (i.e., Lagrange multipliers in Equation~\eqref{eq:f1_min}). 
For example, suppose we have $m$ classes 
$C_t \in \mathbb{C} = \{C^1, ..., C^m\}$ in the data set. Then each distinct 
consecutive pair of classes 
$(C_{t-1},C_t)$
has its own weight vector 
$\mathbf{\theta}^b(C_{t-1},C_t)$, for a total number of pairs and weight
vectors $m^2$. 
An optional {\it START} tag can be added in front of a sequence, forming $m$ 
additional pairs 
$({\it START},C_1)$ (and corresponding weight vectors). 
Besides these pair vectors that accommodate bigram features, each class also 
has a separate weight vector 
$\mathbf{\theta}^u(C_t)$ for unigram features in our model. So in total, we 
use $m$ unigram feature vectors, and $m^2(+m)$ bigram feature vectors to capture the linear chain information.

From Equation~\eqref{eq:f1_max}, we can see that the Lagrange potentials $\psi(\check{\bf y})$ are related to the choice of $\check{\bf y}$ only, and because the binary nature of the F-score of a specific target class~\citep{jurafsky2008speech}, the {\it GFM} algorithm~\citep{dembczynski2011exact} can be applied directly to the  binarized sequences for the target. 

For the adversary's best response, we rewrite Equation~\eqref{eq:f1_min} for a particular target class $C$ by considering the total number of target class in $\hat{\bf y}$ and $\check{\bf y}$ sequence as $\hat{k}$ and $\check{k}$ respectively, as follows:
{\footnotesize
\begin{align}
\check{{\bf y}}^* &= \argmin_{\check{{\bf y}}^{(\check{k})^*}} {\sum_{\hat{{\bf y}}}}p(\hat{{\bf y}})\bigg\{\frac{2\sum_{t=1}^{n}\hat{y}_{Ct}\check{y}_{Ct}}{\hat{k} + \check{k}} 
\notag \\
& \qquad - \sum_{t=1}^{n}\left[\psi_{t}^u(\check{y}_t)+\psi_{t}^b(\check{y}_{t-1},\check{y}_t)\right]\bigg\}, \notag \\
\check{{\bf y}}^{(\check{k})^*} &= \argmin_{\check{{\bf y}} \in \mathbb{\check{Y}}^{(\check{k})}} {\sum_{t=1}^{n}\Bigg\{\left(\sum_{\hat{{\bf y}}}p(\hat{{\bf y}})\frac{2\hat{y}_{Ct}\check{y}_{Ct}}{\hat{k} + \check{k}}\right)} 
\notag \\
& \qquad\qquad 
- \left[\psi_{t}^u(\check{y}_t)+\psi_{t}^b(\check{y}_{t-1},\check{y}_t)\right]\Bigg\} \nonumber 
\end{align}
\begin{align}
= \argmax_{\check{{\bf y}} \in \mathbb{\check{Y}}^{(\check{k})}} \sum_{t=1}^{n}\bigg\{& \psi_{t}^u(\check{y}_t)+\psi_{t}^b(\check{y}_{t-1},\check{y}_t) 
- \underbrace{{\sum_{\hat{k}=1}^{n}}\frac{2p^{C}_{t\hat{k}}}{\hat{k} + \check{k}}}_{f_{t\check{k}}} \check{y}_{Ct} \bigg\}, \label{eq:linear_chain_f1_k}
\end{align}
}%

where $\hat{y}_{Ct}=I(\hat{y}_t=C)$, $\check{y}_{Ct}=I(\check{y}_t=C)$, $\hat{k} = \sum_{t=1}^{n}\hat{y}_{Ct}$, $\check{k} = \sum_{t=1}^{n}\check{y}_{Ct}$ and $p^{C}_{t\hat{k}}$ is the marginal probability that $|\hat{y}_C| = \hat{k}$ and $\hat{y}_t = C$.

The difficulty of solving Eq.~\eqref{eq:linear_chain_f1_k} comes from the Lagrange potential term $[\psi_{t}^u(\check{y}_t)+\psi_{t}^b(\check{y}_{t-1},\check{y}_t)]$ in the linear-chain structure. 
To solve this problem efficiently, we propose a {dynamic programming} algorithm, for the particular class $C$, {\it Linear-Chain F-score Minimizer (LCFM)} in Algorithm~\ref{alg:linear_chain_f1}. 
The MSUM subroutine computes the sum in Equation \eqref{eq:linear_chain_f1_k} via a backward pass with the following recurrence relation for $\check{k}$ instances of class $C$ tags in sequence $\check{\bf y}$:
{\small
\begin{align}
    \beta^{\check{k}}(c_{t-1},t,r) =&
    \begin{cases}
    0 & \!\!t=0\\
    \max_{c_t}\{\beta^{\check{k}}(c_t,t+1,r-I(c_t = C)) 
    & \!\!t > 0\\ \nonumber
     + \psi(c_{t-1},c_t) - f_{t\check{k}} \times I(c_t = C) \}. 
    \end{cases}
\end{align}
}
Looping through subroutines of {\it MSUM} for each of the $n$ values of $\check{k}$, i.e., number of target tags in the best response sequence, can be accomplished 
in $\mathcal{O}(m^2n^3)$ time, which characterizes the overall complexity of the algorithm.

\paragraph{Approximate Best Responses:}
To overcome this cubic complexity, we employ an approximation method for F-score over the linear-chain structure using the cost sensitive approach proposed by \cite{Parambath2014}. 
For different costs of false negatives chosen as $\delta_{+} \in \{2-w, w \in \{0.1,0.2,\dots,1\}\} $ against false positive cost of $ \delta_{-} \in \{w, w \in \{0.1,0.2,\dots,1\}\}$, we find the best response over the linear chain by a Viterbi forward pass in $O(nm^2)$, i.e for the adversary we compute:
{\small
\begin{align} \label{eq:cost_sensitive_approx}
    \alpha^{w}(c_t) =& \begin{cases}
     0  & \!\!\!\!\!\!\!\! t = 0\\
     \max_{c_{t-1}}\{\alpha^{w}(c_{t-1}) + \psi(c_{t-1},c_t)\} & \!\!\!\!\!\!\!\! t>0 \\ \; +I(c_t = C)(1-P^C_t)\delta^w_{-}  + I(c_t \neq C)P^C_t\delta^w_{+}   
   \end{cases}
\end{align}
}%
where $P^C_t$ is the marginal probability that $\check{\mathbf{y}}_t = C$.

The algorithm finds the approximated best response by calling the dynamic programming subroutine for different values of $w$ and chooses the sequence with the lowest F-score \textit{a posteriori}. For the maximizer player, in absence of Lagrangian potentials, choosing the best tag at each position becomes an independent binary decision (target vs. non-target tag) which can be accomplished in $O(n)$. We transform the range of Lagrangian potentials in \eqref{eq:cost_sensitive_approx} to $[-1,1]$ in order to better match the range of cost terms.

\section{EXPERIMENTS AND RESULTS}
\label{sec:experiments_and_results}
We evaluate our approach against 
a {\it maximum margin structured prediction model / SSVM}~\citep{taskar2005discriminative, tsochantaridis2004support} for alignment error rate
and {\it conditional random field (CRF)} for linear-chain F-score. 
Since the maximum margin method's implementation is not available, we implemented it ourselves following the algorithm description~\citep{taskar2005learning,taskar2005discriminative,taskar2005structured}.
We use the {\it Stanford Named Entity Recognizer}'s {\it CRF} implementation~\citep{finkel2005incorporating} in our experiments.

We use the NAACL 2003 Hansards data~\citep{mihalcea2003evaluation} for the 
AER task.
It contains 1,470,000 unlabeled sentence pairs, 447 labeled pairs, and a
separate validation set of 37 labeled pairs.
We experiment with translation from English to French, following the same setting as \cite{taskar2005discriminative} and \cite{cherry2006soft}.  We use first 100 English-French sentence pairs from the original labeled data as training examples, the remaining 347 sentence pairs as test examples, and the same 37 validation pairs as validation examples.
Since the features described in~\cite{taskar2005discriminative} are not available, 
we duplicate them with our best efforts.\footnote{Differences in SSVM performance from \citep{taskar2005discriminative} suggest that our features differ from those used previously.} 

We train the maximum margin structured model ($SSVM$), and our adversarial
approach (ADV) with maximum weight bipartite matching ($ADV_{bip}$), and without bipartite matching constraints for AER ($ADV_{aer}$) using those features extracted from the training data set. 
Also, following the same setting as~\cite{taskar2005discriminative}, we include GIZA++'s unsupervised prediction from the $1^5H^53^34^3$ training scheme \citep{och2003systematic} as an additional feature.
We select $\ell_2$ regularization values using 
performance on the validation data set. 

\begin{table}[htb]
\centering
\small
\caption{AER of different models.}
\label{tab:aer_perf}
\begin{tabular}{|l|c|c|}
\hline
\textbf{Model} & \textbf{Valid} & \textbf{Test} \\ \hline
$SSVM
$ & 13.98\% & 13.34\% \\ 
$MPG_{bip}
$ & 15.82\% & 14.52\% \\ 
$MPG_{aer}
$ & \textbf{6.97}\% & \textbf{7.28\%} \\ \hline
\end{tabular}
\end{table}

\begin{table}[htb]
\centering

\caption{\footnotesize $F_1$ scores of $CRF$ and $MPG_{f_1}$ on `testa' (top) and `testb' (bottom).}
\label{tab:f1}
\normalsize
\begin{tabular}{|c|l|c|c|c|}
\hline
\multicolumn{2}{|l|}{\multirow{2}{*}{Dataset}} & \multicolumn{3}{c|}{\textbf{testa}}  
\\ \cline{3-5} 
\multicolumn{2}{|l|}{} & \multicolumn{1}{c|}{$CRF$} & \multicolumn{1}{c|}{$ADV_{f_1}$} & \multicolumn{1}{c|}{$ADV^{\approx}_{f_1}$} 
\\ \hline
\multirow{4}{*}{\rotatebox{90}{\textbf{conll$_{300}$}}} & PER & 17.88 & \textbf{29.64} & \textbf{29.64}
\\ 
 & LOC & 30.94 & \textbf{47.40} & \textbf{47.40}
 \\
 & ORG & 7.19 & \textbf{15.74} & \textbf{15.74} 
 \\
 & MISC & 16.22 & \textbf{19.51} & \textbf{19.51} 
 \\ \hline
\multirow{4}{*}{\begin{turn}{90}\textbf{conll$_{1000}$}\end{turn}} & PER & 78.68 & \textbf{85.42} & \textbf{85.42}
\\
 & LOC & 80.00 & 78.63  & 78.63
 \\
 & ORG & 58.24 & \textbf{59.39} & \textbf{59.39} 
 \\
 & MISC & 62.30 & \textbf{67.08} & \textbf{67.08} 
 \\ \hline
\multirow{4}{*}{\begin{turn}{90}\textbf{conll$_{3000}$}\end{turn}} 
& PER & 85.72 & \textbf{88.97} & \textbf{88.97} 
\\
 & LOC & 85.74 & 85.38 & 85.38 
 \\
 & ORG & 75.69 & \textbf{76.34} & \textbf{76.34} 
 \\
 & MISC & 78.82 & \textbf{81.04} & \textbf{81.04} 
 \\ \hline
\end{tabular}

\vspace{3mm}

\begin{tabular}{|c|l|c|c|c|}
\hline
\multicolumn{2}{|l|}{\multirow{2}{*}{Dataset}} & 
\multicolumn{3}{c|}{\textbf{testb}} \\ \cline{3-5} 
\multicolumn{2}{|l|}{} & 
\multicolumn{1}{c|}{$CRF$} & \multicolumn{1}{c|}{$ADV_{f_1}$} & \multicolumn{1}{c|}{$ADV^{\approx}_{f_1}$} 
\\ \hline
\multirow{4}{*}{\rotatebox{90}{\textbf{conll$_{300}$}}} & PER & 
40.00 & \textbf{44.68} & \textbf{44.68}\\ 
 & LOC & 
 52.10 & \textbf{66.34} & \textbf{66.34}\\
 & ORG 
 & 14.16 & \textbf{19.04} & \textbf{19.04} \\
 & MISC & 
 30.43 & \textbf{35.21} & \textbf{35.21} \\ \hline
\multirow{4}{*}{\begin{turn}{90}\textbf{conll$_{1000}$}\end{turn}} & PER & 
76.36 &  \textbf{83.21} & \textbf{83.21}\\
 & LOC & 
 76.65 & \textbf{76.81} & \textbf{76.81}\\
 & ORG & 
 61.14 & \textbf{61.73} & \textbf{61.73}\\
 & MISC & 
 59.28 & \textbf{66.06} & \textbf{66.06} \\ \hline
\multirow{4}{*}{\begin{turn}{90}\textbf{conll$_{3000}$}\end{turn}} & PER & 
81.05 & \textbf{84.93} & \textbf{84.93} \\
 & LOC 
 & 82.80 &  80.82 & 80.82 \\
 & ORG 
 & 66.18 & \textbf{69.57} & \textbf{69.57}\\
 & MISC 
 & 70.05 & \textbf{75.32} & \textbf{75.32} \\ \hline
\end{tabular}

\centering

\end{table}

The performances of models on validation and test data sets are shown in Table~\ref{tab:aer_perf}.
$ADV_{aer}$ outperforms all other model, 
which demonstrates the effectiveness of better aligning the predictor to the performance measure of interest.
Comparing $ADV_{bip}$ against $ADV_{aer}$ shows the advantage of modeling AER more directly without as strong exclusivity restrictions on the edges over modeling the alignment problem as a maximum weight bipartite matching.

In the comparison to the CRF model, we use the well known CoNLL-2003 English data set~\citep{tjong2003introduction}.
We consider each sentence as one sequence example, 
and create three different sizes of subsets from the CoNLL-2003 data for our experiments, 
to demonstrate the benefit that adversarially optimizing $F_1$ can bring, with respect to the training data size.
The first data set contains the first 300 sentences from `train,' 300 sentences from `testa,' and 300 sentences from `testb.' 
The second data set contains 1000 ($\times$3) sentences, and the last contains 3000 ($\times$3) sentences.
Features are extracted exact the same as Stanford NER with the first-order CRF configuration.
The number of features in each data set is 31979, 67513, and 166737 respectively. 

We evaluate the $F_1$ score of $ADV_{f_1}$ on each data set. 
The performances of both the CRF model and the $ADV_{f_1}$ model can be found in Table~\ref{tab:f1}.
$ADV_{f_1}$ works better for all NER tags than CRF on the 300 sentences data sets.
On 1000 and 3000 sentences data sets, $ADV_{f_1}$ achieves better $F_1$ scores for all tags except 'LOC', for which CRF shows better results. 
The results suggest that optimizing $F_1$ directly can reduce the need for using larger data set.

\begin{figure}[htb]
\centering
\includegraphics[width=0.47\textwidth]{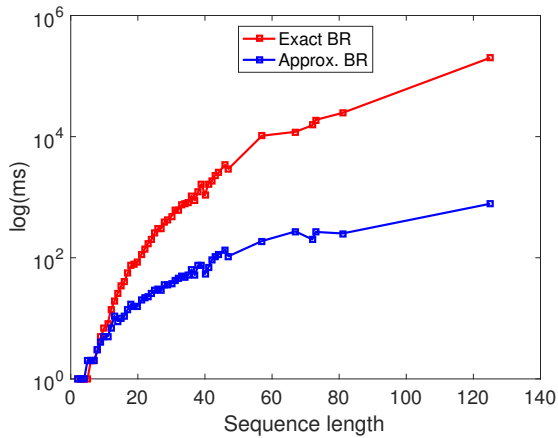}
\caption{Comparison of double oracle performance using exact best response (LCFM Algorithm \ref{alg:linear_chain_f1}) and the alternative approximation method on different lengths of the sequence.}
\vspace{-2mm}
\label{fig:br_comparison}
\end{figure}

We train $ADV^{\approx}_{f_1}$ using our faster approximation method for best response (\S\ref{sec:chainF}).  This facilitates more efficient training by an order of magnitude. 
Solving the game by this approximation yields same best single strategy as using exact method in Algorithm \ref{alg:linear_chain_f1}, $92\% \pm .3$ of the time with 99\% confidence level. Figure~\ref{fig:br_comparison} shows the efficiency of this method for solving the game for longer sequences using double oracle. We still use the exact method for prediction at test time. The results in Table~\ref{tab:f1} suggest no penalty in final test scores. 

\section{CONCLUSION \& FUTURE WORK}

This paper generalized adversarial prediction methods to structured prediction tasks with multivariate performance measures.
We investigated the benefits of this approach by
addressing two key NLP tasks:  word alignment evaluated using the alignment error rate and named entity recognition using chain structures evaluated using the F-score.
The algorithms for finding the best response for each task are described in detail.
In our future work, we plan to further extend adversarial prediction  to other 
tasks of interest in NLP and computer vision.
Also, we plan to further study and characterize the multivariate performance measures that can be efficiently optimized within the adversarial prediction  framework, and explore the effectiveness of approximation during constraint generation.

\bibliography{mpg_aistats_2018}
\bibliographystyle{abbrvnat}

\end{document}